\documentclass[journal,onecolumn,11pt,draftclsnofoot]{IEEEtran}
\usepackage{times,amsmath,amssymb,amsfonts,amsthm}
\usepackage{graphicx}
\usepackage{fancybox}
\usepackage{subfigure}
\usepackage{algorithm}
\usepackage{algorithmic}
\usepackage{xcolor}
\usepackage{hyperref}
\usepackage{url}

\newcommand{\R}{\mathbb{R}}

\newcommand{\matA}{\boldsymbol{A}}
\newcommand{\matB}{\boldsymbol{B}}
\newcommand{\matC}{\boldsymbol{C}}

\newcommand{\vetx}{\boldsymbol{x}}
\newcommand{\vety}{\boldsymbol{y}}

\newcommand{\bb}{\begin{equation}}
\newcommand{\ee}{\end{equation}}
\newcommand{\bbb}{\begin{eqnarray}}
\newcommand{\eee}{\end{eqnarray}}
\newcommand{\benu}{\begin{enumerate}}
\newcommand{\eenu}{\end{enumerate}}

\newcommand{\bpm}{\begin{bmatrix}}
\newcommand{\epm}{\end{bmatrix}}

\newcommand{\ii}{\boldsymbol{i}}
\newcommand{\jj}{\boldsymbol{j}}
\newcommand{\kk}{\boldsymbol{k}}

\newcommand{\quat}[1]{{#1}_0 + {#1}_1 \ii + {#1}_2 \jj + {#1}_3 \kk}

\newcommand{\cossec}{\operatorname{\cossec}} \def\cossec{cossec}

\newcommand{\qeq}{\quad \mbox{and} \quad}

\newcommand{\basis}{\mathcal{E}=\{e_1,\ldots,e_n\}}

\newtheorem{definition}{Definition}
\newtheorem{theorem}{Theorem}

\newtheorem{proposition}{Proposition}

\theoremstyle{definition}
\newtheorem{example}{Example}
\newtheorem{remark}{Remark}

\begin{document}
%
\title{Understanding Vector-Valued Neural Networks and Their Relationship with Real and Hypercomplex-Valued Neural Networks
\thanks{This work has been partially supported by the National Council for Scientiﬁc and Technological Development (CNPq) under grants 315820/2021-7 and São Paulo Research Foundation (FAPESP) under grant 2022/01831-2.}
}
%
%
%

\author{Marcos Eduardo Valle,~\IEEEmembership{Member,~IEEE}
\thanks{Marcos Eduardo Valle is with the Universidade Estadual de Campinas (UNICAMP), Campinas -- Brazil, e-mail: valle@ime.unicamp.br.}
}

\maketitle

\begin{abstract}
Despite the many successful applications of deep learning models for multidimensional signal and image processing, most traditional neural networks process data represented by (multidimensional) arrays of real numbers. The intercorrelation between feature channels is usually expected to be learned from the training data, requiring numerous parameters and careful training. In contrast, vector-valued neural networks are conceived to process arrays of vectors and naturally consider the intercorrelation between feature channels. Consequently, they usually have fewer parameters and often undergo more robust training than traditional neural networks. This paper aims to present a broad framework for vector-valued neural networks, referred to as V-nets. In this context, hypercomplex-valued neural networks are regarded as vector-valued models with additional algebraic properties. Furthermore, this paper explains the relationship between vector-valued and traditional neural networks. Precisely, a vector-valued neural network can be obtained by placing restrictions on a real-valued model to consider the intercorrelation between feature channels. Finally, we show how V-nets, including hypercomplex-valued neural networks, can be implemented in current deep-learning libraries as real-valued networks. 
\end{abstract}

\begin{IEEEkeywords}
\noindent Multidimensional signal and image processing, vector-valued neural network, hypercomplex-valued neural network, deep learning.
\end{IEEEkeywords}

%

\section{Introduction}
\noindent Neural networks achieved outstanding performance in many signal and image processing tasks, especially with the advent of the deep learning paradigm \cite{Lecun2015DeepLearning,Gonzalez2018DeepNetworks}. Despite their many successful applications, traditional neural networks are theoretically designed to process real-valued or, at most, complex-valued data. Accordingly, signals and images are represented by (possibly multidimensional) arrays of real or complex numbers \cite{Lecun2015DeepLearning,Gonzalez2018DeepNetworks,Geron19HandsOn}. {Furthermore, most traditional neural networks do not consider possible intercorrelation between feature channels beforehand. Indeed, although strategies like the squeeze-and-excitation block boost interdependencies between feature channels \cite{Hu2017Squeeze-and-ExcitationNetworks}, such relationships are expected to be learned from the training data.} Consequently, besides relying on appropriate loss functions and effective optimizers, traditional deep learning models usually have too many parameters and demand a long training time. 

In contrast, vector-valued neural networks (V-nets) are designed to process arrays of vectors. They naturally take into account the intercorrelation between feature channels {a priori}. Hence, V-nets are expected to have fewer parameters than traditional neural networks. Furthermore, they should be less susceptible to being trapped in a local minimum of the loss function surface during the training. Hypercomplex-valued neural networks are examples of robust and lightweight V-nets {that, besides dealing with vector-valued data, can take geometrical concepts into account \cite{Vieira2022AMachines,Vieira2022AcuteNetworks,Grassucci2022PHNNs:Convolutions,Grassucci2023DualRepresentation,Buchholz2000QuaternionicMLP,Ruhe2023GeometricNetworks}.} 

This paper aims to present a detailed framework for V-nets, making plain and understandable their relationship with traditional and hypercomplex-valued neural networks. Precisely, we first present the mathematical background for vector-valued neural networks. {Then, we address their relationship with traditional and hypercomplex-valued neural networks, focusing on dense and convolutional layers.} On the one hand, hypercomplex-valued neural networks are regarded as vector-valued models with additional algebraic properties. On the other hand, V-nets can be viewed as traditional neural networks with restrictions to take into account the intercorrelation between the feature channels. Using these relationships, we show how to emulate vector-valued (and hypercomplex-valued) neural networks using traditional models, allowing us to implement them using current deep-learning libraries. 

The paper is structured as follows. Section \ref{sec:background} provides the mathematical background for V-nets, including hypercomplex algebras and examples. Basic vector-valued matrix operations and their relationship with traditional linear algebra are briefly reviewed in Section \ref{sec:matrix}.
Section \ref{sec:VvNNs} introduces V-nets, with a focus on dense and convolutional layers. This section also addresses the approximation capability of shallow dense networks and explains how to implement V-nets using the current deep-learning libraries designed for real-valued data. The paper finishes with concluding remarks in Section \ref{sec:concluding}.

\section{Vector and Hypercomplex Algebras} \label{sec:background}

Despite their many successful applications, traditional neural networks are designed to process arrays of real numbers. However, many image and signal-processing tasks -- such as those related to multivariate images and 3D audio signals \cite{Grassucci2023DualRepresentation,Parcollet2020ANetworks,Sebastian2023Quaternionsoverview} -- are concerned with vector-valued data, which can be better explored by considering models designed to deal with arrays of vectors. Because addition and multiplication are core operations for designing effective neural networks, this section reviews some key concepts of algebra. Broadly speaking, an algebra is a vector space (with component-wise vector addition) enriched with a multiplication of vectors. Such mathematical structure yields the background for developing vector-valued and hypercomplex-valued neural networks.  

\begin{definition}[Algebra \cite{Schafer1961AnAlgebras}] \label{def:algebra}
 An algebra $\mathbb{V}$ is a vector space over a field $\mathbb{F}$ with an additional bilinear operation called multiplication or product.    
\end{definition}

As a bilinear operation, the multiplication of $x,y \in \mathbb{V}$, denoted by the juxtaposition $xy$, satisfies 
 \begin{equation*}
  (x+y)z = xz + yz \qeq z(x+y)=zx+zy, \quad \forall x,y,z \in \mathbb{V},
 \end{equation*}
 and 
 \begin{equation*}
  \alpha(xy) = (\alpha x) y = x (\alpha y), \quad \forall \alpha \in \mathbb{F} \qeq x,y \in \mathbb{V}.
 \end{equation*}

\begin{remark}
Because we are mainly concerned with the implementation of models on traditional computers, for the sake of simplicity, we only consider algebras over the field of real numbers, that is, $\mathbb{F}=\mathbb{R}$. Furthermore, we will only be concerned with finite dimensional vector spaces. In other words, we assume that $\mathbb{V}$ is a vector space of dimension $n$, i.e., $dim(\mathbb{V})=n$.
\end{remark}

Let $\mathcal{E} = \{e_1,e_2,\ldots,e_n\}$ be an ordered basis for $\mathbb{V}$. Given $x \in \mathbb{V}$, there is an unique $n$-tuple $(x_1,x_2,\ldots,x_n) \in \mathbb{R}^n$ such that 
\begin{equation*}
 x = \sum_{i=1}^n x_i e_i.
\end{equation*}
The scalars $x_1,\ldots,x_n$ are the coordinates of $x$ relative to the ordered basis $\mathcal{E}$. In computational applications, $x \in \mathbb{V}$ is given by its coordinates relative to the ordered basis $\basis$. Precisely, $x$ is usually given by a vector in $\R^n$, and the canonical basis is often implicitly considered. In order to further distinguish $x \in \mathbb{V}$ from the $n$-tuple $(x_1,\ldots,x_n) \in \mathbb{R}^n$, we introduce the following isomorphism: 
\begin{definition}[Isomorphism between $\mathbb{V}$ and $\mathbb{R}^n$] \label{def:isomorphism}
Given an ordered basis $\basis$, the mapping $\varphi:\mathbb{V}\to \R^n$ given by 
\begin{equation}
 \label{eq:varphi}
 \varphi(x) = \begin{bmatrix} x_1 \\ \vdots \\ x_n\end{bmatrix} \in \R^n, \quad \forall x \in \mathbb{V},
\end{equation}
yields an isomorphism between $\mathbb{V}$ and $\mathbb{R}^n$.  
\end{definition}

Using the isomorphism $\varphi$, $\mathbb{V}$ inherits the topology and metric from $\mathbb{R}^n$. For example, we can define the absolute value of $x \in \mathbb{V}$ with respect to the basis $\basis$ as the Euclidean norm of $\varphi(x)$:
\begin{equation}
 \label{eq:abs}
 |x|:= \|\varphi(x)\|_2 = \sqrt{x_1^2+x_2^2+\ldots+x_n^2}.
\end{equation}
We would like to remark, however, that the absolute value of $x$ given by \eqref{eq:abs} is not an invariant; it depends on the basis $\basis$. Like traditional linear algebra, the basis $\mathcal{E}$ plays a crucial role in the algebra $\mathbb{V}$. 

Let us now show how multiplication is defined on $\mathbb{V}$. Given an ordered basis $\mathcal{E}=\{e_1,\ldots,e_n\}$, the multiplication is completely determined by the $n^3$ parameters $p_{ijk} \in \mathbb{R}$ which appear in the products
\begin{equation}
 \label{eq:multiplication_table}
 e_i e_j = \sum_{k=1}^n p_{ijk} e_k, \quad \forall i,j =1,\ldots,n.
\end{equation}
The products in \eqref{eq:multiplication_table} can be arranged in the so-called multiplication table:
\begin{center}
 \begin{tabular}{c|ccc}
   &  & $e_j$ & \\ \hline 
    & & $\vdots$ &  \\
   $e_i$ & $\cdots$ & $\sum_{k=1}^n p_{ijk}e_k$ & $\cdots$ \\
   & & $\vdots$ &  
 \end{tabular}
\end{center}

The properties of an algebra can be obtained by analyzing the basis elements or the multiplication table. For example, an algebra $\mathbb{V}$ is considered commutative if 
 \begin{equation*}
  xy = yx, \quad \forall x,y \in \mathbb{V}.
 \end{equation*}
 Given an ordered basis $\basis$, the algebra is commutative if and only if 
 \begin{equation*}
     e_i e_j = e_j e_i, \quad \forall i,j =1,\ldots,n.
 \end{equation*}
Equivalently, from the multiplication table, we conclude that an algebra is commutative if and only if 
   \begin{equation*}
   p_{ijk} = p_{jik}, \quad \forall i,j,k=1,\ldots,n.
   \end{equation*}
Analogously, an algebra $\mathbb{V}$ is associative if 
 \begin{equation*}
  (xy)z = x(yz), \quad \forall x,y,z \in \mathbb{V}.
 \end{equation*}
 Thus, the algebra is associative if and only if 
  \begin{equation*} (e_i e_j) e_k = e_i (e_j e_k), \quad \forall i,j,k =1,\ldots,n.
  \end{equation*} 
In other words, an algebra is associative if and only if 
   \begin{equation*} \sum_{\mu=1}^n p_{ij\mu} p_{k\mu\ell} = \sum_{\mu=1}^n p_{jk\mu}p_{i\mu\ell}, \quad \forall i,j,k,\ell=1,\ldots,n.
   \end{equation*}

The interest in machine learning techniques and neural network models based on hypercomplex algebras, including predominantly complex numbers and quaternions, has a long history \cite{Parcollet2020ANetworks,Sebastian2023Quaternionsoverview,aizenberg11book,hirose12,Lee2022Complex-ValuedSurvey}. Many researchers (including myself) list the capability to treat multidimensional data as a single entity as one prominent advantage of hypercomplex-valued models. According to Definition \ref{def:algebra}, however, any algebra provides the mathematical background for dealing with arrays of vectors. Therefore, I suggest defining hypercomplex algebras as algebra with additional geometric or algebraic properties. Precisely, I propose the following:
\begin{definition}[Hypercomplex algebra \cite{Kantor1989HypercomplexAlgebras,Catoni2008TheSpace-Time}] \label{def:hypercomplex}
A hypercomplex algebra, denoted by $\mathbb{H}$, is a finite-dimensional algebra in which the product has a two-sided identity.
\end{definition}

From Definition \ref{def:hypercomplex}, a hypercomplex algebra $\mathbb{H}$ is equipped with an unique element $e_0$ such that 
\begin{equation*}
xe_0 = e_0x = x, \quad \forall x \in \mathbb{H}.
\end{equation*}
The identity is usually the first element of the ordered basis. Thus, $\mathcal{E}=\{e_0,e_1,\ldots,e_n\}$ is an ordered basis of an hypercomplex algebra and $dim(\mathbb{H})=n+1$. Moreover, we often consider the canonical basis $\tau = \{1,\ii_1,\ldots,\ii_n\}$. Accordingly, a hypercomplex number is given by 
\begin{equation*} 
 x = x_0+x_1\ii_1+\ldots+x_n\ii_n.
\end{equation*}
The multiplication table of a hypercomplex algebra with respect to the canonical basis $\tau = \{1,\ii_1,\ldots,\ii_n\}$ is
 \begin{center}
 \begin{tabular}{c|cccc}
   & $1$ & $\ii_1$ & $\ii_j$ & $\ii_n$ \\ \hline 
   $1$ & $1$ & $\ii_1$ & $\ii_j$ & $\ii_n$ \\
   & & & $\vdots$ &  \\
   $\ii_i$ & $\ii_i$ & $\cdots$ & $p_{ij0}+\sum_{k=1}^n p_{ijk}\ii_k$ & $\cdots$ \\
   & & & $\vdots$ &  
 \end{tabular}
\end{center}

\begin{remark}
We would like to remark that Definition \ref{def:hypercomplex} is consistent with the general approach of Kantor and Solodovnik and includes well-known hypercomplex algebras as particular instances \cite{Kantor1989HypercomplexAlgebras}. In particular, all Clifford and Cayley-Dickson algebras are examples of hypercomplex algebras. 
\end{remark}

Let us return our attention to an arbitrary finite-dimensional algebra $\mathbb{V}$. Using the distributive law and the multiplication table, the product of $x = \sum_{i=1}^n x_i e_i$ and $y=\sum_{j=1}^n y_j e_j$ satisfies
 \begin{align*}
  xy &= \left(\sum_{i=1}^n x_i e_i\right)\left(\sum_{j=1}^n y_j e_j\right) 
  = \sum_{i=1}^n \sum_{j=1}^n x_i y_j (e_i e_j)  
  = \sum_{k=1}^n \left(\sum_{i=1}^n \sum_{j=1}^n x_i y_j p_{ijk} \right) e_k.
 \end{align*}
Because the product is bilinear, the function $\mathcal{B}_k:\mathbb{V}\times \mathbb{V} \to \mathbb{R}$ given by
 \begin{equation*}
 \mathcal{B}_k(x,y) = \sum_{i=1}^n \sum_{j=1}^n x_i y_j p_{ijk}, \quad \forall k=1,\ldots,n, 
 \end{equation*}
 is a bilinear form. Therefore, we obtain the following proposition \cite{Vital2022ExtendingNetworks}:
\begin{proposition} \label{prop:bilinear}
  Let $\mathcal{E}=\{e_1,\ldots,e_n\}$ be an ordered basis of an algebra $\mathbb{V}$. The multiplication of $x=\sum_{i=1}^n x_i e_i$ and $y=\sum_{j=1}^n y_j e_j$ satisfies
  \begin{equation} \label{eq:prod_bilinear}
   xy = \sum_{k=1}^n \mathcal{B}_k(x,y) e_k,
  \end{equation}
  where $\mathcal{B}_k:\mathbb{V}\times \mathbb{V} \to \mathbb{R}$ is a bilinear form whose matrix representation in the ordered basis $\mathcal{E}$ is 
  \begin{equation*}
   \boldsymbol{B}_k = \begin{bmatrix}
p_{11k} & p_{12k} & \ldots & p_{1nk} \\
p_{21k} & p_{22k} & \ldots & p_{2nk} \\ 
  \vdots & \vdots & \ddots & \vdots \\
p_{n1k} & p_{n2k} & \ldots & p_{nnk}  
                       \end{bmatrix} \in \R^{n \times n}, \quad \forall k=1,\ldots,n.
  \end{equation*}
 Thus, we have $\mathcal{B}_k(x,y) = \varphi(x)^T \boldsymbol{B}_k \varphi(y)$.
 \end{proposition}

Using Proposition \ref{prop:bilinear}, we introduce the following definition that plays an important role in the approximation capability of V-nets such as the vector-valued multilayer perceptron (V-MLP) network \cite{Vital2022ExtendingNetworks}:
\begin{definition}[Non-degenerate algebra]
An algebra $\mathbb{V}$ is non-degenerate respect to an ordered basis $\basis$ if all the bilinear forms $\mathcal{B}_1,\ldots,\mathcal{B}_n$ in \eqref{eq:prod_bilinear} are non-degenerate. Otherwise, we say that the algebra $\mathbb{V}$ is degenerate with respect to $\mathcal{E}$. 
\end{definition}

Recall that a bilinear form $\mathcal{B}_k:\mathbb{V}\times \mathbb{V} \to \R$ is non-degenerate if its matrix representation $\boldsymbol{B}_k$ is non-singular.

In addition to expressing the multiplication of two vectors through bilinear forms, it can also be represented as a matrix-vector operation. Precisely, the multiplication to the left by a vector $a = \sum_{i=1}^n a_i e_i \in \mathbb{V}$ yields a linear operator $\mathcal{A}_L: \mathbb{V} \to \mathbb{V}$ defined by $\mathcal{A}_L (x) = ax$, for all $x\in\mathbb{V}$. Therefore, the matrix representation of $\mathcal{A}_L$ relative to an ordered basis $\basis$ yields a mapping $\mathcal{M}_L:\mathbb{V} \to \R^{n \times n}$ given by
\begin{equation}
 \mathcal{M}_L(a) = 
 \begin{bmatrix}
| & | & & | \\
\varphi(ae_1) & \varphi(ae_2) & \ldots & \varphi(ae_n) \\
| & | & & | 
\end{bmatrix} \nonumber =
 \begin{bmatrix}
\sum_{i=1}^n a_i p_{i11} & \sum_{i=1}^n a_i p_{i21} & \ldots & \sum_{i=1}^n a_i p_{in1} \\
\sum_{i=1}^n a_i p_{i12} & \sum_{i=1}^n a_i p_{i22} & \ldots & \sum_{i=1}^n a_i p_{in2} \\ 
\vdots & \vdots & \ddots & \vdots \\
\sum_{i=1}^n a_i p_{i1n} & \sum_{i=1}^n a_i p_{i2n} & \ldots & \sum_{i=1}^n a_i p_{inn} 
\end{bmatrix}.
\end{equation}
In words, $\mathcal{M}_L:\mathbb{V} \to \mathbb{R}^{n \times n}$ maps a vector $a \in \mathbb{V}$ to its matrix representation in the multiplication by the left with respect to the ordered basis $\mathcal{E}$. Alternatively, we can write
 \begin{equation}
  \label{eq:matrix_AL}
  \mathcal{M}_L(a)
  = \sum_{i=1}^n a_i \boldsymbol{P}_{i:}^T, \quad \text{with} \quad
  \boldsymbol{P}_{i:}^T = \begin{bmatrix}
     p_{i11} & p_{i21} & \ldots & p_{in1} \\
     p_{i12} & p_{i22} & \ldots & p_{in2} \\
     \vdots & \vdots & \ddots & \vdots \\
     p_{i1n} & p_{i2n} & \ldots & p_{inn} \end{bmatrix}.
 \end{equation} 
Using the matrix representation, we have 
\begin{equation} 
\label{eq:ax}
 \varphi(ax) = \mathcal{M}_L(a)\varphi(x) = \sum_{i=1}^n a_i \boldsymbol{P}_{i:}^T \varphi(x),
\end{equation} 
for all $a=\sum_{i=1}^n a_i e_i \in \mathbb{V}$ and $x \in \mathbb{V}$. Note that \eqref{eq:ax} provides an efficient formula for computing vector multiplication using traditional matrix operations. 


\begin{example}[Quaternions] \label{ex:quaternions}
Consider the quaternions with the canonical basis $\tau = \{1,\ii,\jj,\kk\}$. 
The product of $x=\quat{x}$ and $y=\quat{y}$ satisfies
\[ \varphi(xy) = \begin{bmatrix}
 x_0 & - x_1 & - x_2 & - x_3 \\ 
 x_1 & x_0 &  - x_3 &  x_2 \\
 x_2 & x_3 & x_0  & -x_1 \\
 x_3 & - x_2 & x_1 & x_0
\end{bmatrix} \begin{bmatrix} y_0 \\ y_1 \\ y_2 \\ y_3 \end{bmatrix} = \mathcal{M}_L(x) \varphi(y).\]
Note that 
\[ {\mathcal{M}_L(x) = x_0 \boldsymbol{P}_{0:}^T + x_1 \boldsymbol{P}_{1:}^T + x_2 \boldsymbol{P}_{2:}^T + x_3 \boldsymbol{P}_{3:}^T}, \]
where $\boldsymbol{P}_{0:} = \mathbf{I}_{4\times4}$ is the identity matrix and
\[
 \boldsymbol{P}_{1:}^T = \begin{bmatrix}
          0 & -1 & 0 & 0\\
          1 & 0 & 0 & 0 \\
          0 & 0 & 0 & -1 \\
          0 & 0 & 1 & 0
         \end{bmatrix}, \quad
\boldsymbol{P}_{2:}^T = \begin{bmatrix}
          0 & 0 & -1 & 0\\
          0 & 0 & 0 & 1 \\
          1 & 0 & 0 & 0 \\
          0 & -1 & 0 & 0
         \end{bmatrix}, \qeq
\boldsymbol{P}_{3:}^T = \begin{bmatrix}
          0 & 0 & 0 & -1\\
          0 & 0 & -1 & 0 \\
          0 & 1 & 0 & 0 \\
          1 & 0 & 0 & 0
         \end{bmatrix}.
\]
\end{example}

\begin{example}[Parametrized ``Hypercomplex'' Algebras] \label{ex:PHA}
Recently, Zhang et al. introduced the so-called \textit{parametrized ``hypercomplex'' algebras} \cite{Grassucci2022PHNNs:Convolutions,Zhang2021BeyondParameters}. A parametrized ``hypercomplex'' algebra is defined as follows using the matrix representation of the multiplication:
Given matrices $\boldsymbol{P}_1,\ldots,\boldsymbol{P}_n \in \R^{n \times n}$ and an ordered basis $\basis$, the product in a parametrized ``hypercomplex'' algebra is defined by 
\begin{equation} \label{eq:Parametrized}
 xy = \varphi^{-1}\Big(\sum_{i=1}^n x_i \boldsymbol{P}_i \varphi(y) \Big),
\end{equation}
for all $x = \sum_{i=1}^n x_i e_i$ and $y=\sum_{i=1}^n y_i e_i$.
Note that \eqref{eq:Parametrized} is equivalent to \eqref{eq:ax}. Therefore, despite being referred to as ``hypercomplex'', the multiplication given by \eqref{eq:Parametrized} does not necessarily have an identity. Thus, a parameterized ``hypercomplex'' algebra may not meet the criteria to be classified as hypercomplex as per the Definition \ref{def:hypercomplex}. Nevertheless, the multiplication defined by \eqref{eq:Parametrized} has been effectively used to learn the algebra of vector-valued neural networks \cite{Grassucci2022PHNNs:Convolutions,Zhang2021BeyondParameters}.
\end{example}

\section{Vector-Valued Matrix Computation} \label{sec:matrix}

Matrix computation is a key concept for developing efficient vector- and hypercomplex-valued network models because some fundamental building blocks, like dense and convolutional layers, compute affine transformations followed by a non-linear activation function. In this section, we present some basic vector-valued matrix computation concepts \cite{Vieira2022AMachines}.

As in the traditional matrix algebra, the product of two vector-valued matrices $\matA \in \mathbb{V}^{M\times L}$ and $\matB \in \mathbb{V}^{L \times N}$ results in a new matrix $\matC \in \mathbb{V}^{M \times N}$ with entries defined by
\begin{equation*} 
c_{ij} = \sum_{\ell=1}^L a_{i\ell} b_{\ell j}, \quad \forall i=1,\ldots,M \qeq j=1,\ldots,N. 
\end{equation*} 
To take advantage of fast scientific computing software, we compute the above operation using real-valued matrix operations as follows. Using the isomorphism $\varphi: \mathbb{V} \to \mathbb{R}^{n}$ and the mapping $\mathcal{M}_L:\mathbb{V} \to \mathbb{R}^{n \times n}$ defined respectively by \eqref{eq:varphi} and \eqref{eq:matrix_AL},
we obtain
\begin{equation*}
    \varphi(c_{ij})   
    = \sum_{\ell=1}^L \varphi\left( a_{i\ell} b_{\ell j} \right)  
    = \sum_{\ell=1}^L \mathcal{M}_L(a_{i\ell})\varphi(b_{\ell j}).
\end{equation*} 
Equivalently, using real-valued matrix operations, we have
\begin{equation} 
    \label{eq:Matrix-vec-prod} 
    \varphi(\matC) = \mathcal{M}_L(\matA) \varphi(\matB), 
\end{equation} 
where $\mathcal{M}_L$ and $\varphi$ are extended as follows for vector-valued matrices:
\begin{equation}  \label{eq:Matrix-Phi_L}
\mathcal{M}_L(\matA) = \begin{bmatrix}
\mathcal{M}_L(a_{11}) & \mathcal{M}_L(a_{12}) & \ldots & \mathcal{M}_L(a_{1L}) \\
\vdots  & \vdots & \ddots & \vdots \\
\mathcal{M}_L(a_{M1}) & \mathcal{M}_L(a_{M2}) & \ldots & \mathcal{M}_L(a_{ML}) 
\end{bmatrix} \in \R^{nM\times nL},
\end{equation} 
and
\begin{equation} \label{eq:Matrix-varphi}
\varphi(\matB) = \begin{bmatrix}
\varphi(b_{11}) & \ldots & \varphi(b_{1N}) \\
\varphi(b_{21}) & \ldots & \varphi(b_{2N}) \\
\vdots & \ddots & \vdots \\
\varphi(b_{L1}) & \ldots & \varphi(b_{LN}) \\
\end{bmatrix} \in \R^{nL\times N}.
\end{equation}
Therefore, reorganizing the elements of $\varphi(\matC)$, we can write 
\begin{equation*} 
\matC = \varphi^{-1} \left(\mathcal{M}_L(\matA) \varphi(\matB) \right), \end{equation*}
which allows the computation of vector-valued matrix products using the real-valued linear algebra often available in scientific computing software. 

To further reduce the computing time, the real-valued matrix $\mathcal{M}_L(\matA) \in \R^{nM \times nL}$ can be computed using the Kronecker product.  The Kronecker product between two real-valued matrices $\boldsymbol{A} = (a_{ij}) \in \R^{N\times M}$ and $\boldsymbol{B} \in \R^{P \times Q}$, denoted by $\boldsymbol{A} \otimes \boldsymbol{B}$, yields the block matrix defined by
\begin{equation*}
      \boldsymbol{A} \otimes \boldsymbol{B} = 
      \begin{bmatrix}
        a_{11} \boldsymbol{B} & a_{12}\boldsymbol{B} & \ldots & a_{1M}\boldsymbol{B} \\
        a_{21} \boldsymbol{B} & a_{22}\boldsymbol{B} & \ldots & a_{2M}\boldsymbol{B} \\
        \vdots & \vdots & \ddots & \vdots \\
        a_{N1} \boldsymbol{B} & a_{N2}\boldsymbol{B} & \ldots & a_{NM}\boldsymbol{B} \\
      \end{bmatrix} \in \R^{NP \times MQ}.
\end{equation*}
Basic properties and some applications of the Kronecker product can be found in \cite{Stenger1968KroneckerOperators,Loan2000TheProduct}.

As per the references \cite{Grassucci2022PHNNs:Convolutions,Zhang2021BeyondParameters}, we employ the Kronecker product to calculate $\mathcal{M}_L(\matA)$ in the following manner.
From \eqref{eq:matrix_AL}, we have
\begin{equation*}
\mathcal{M}_L(\matA) 
= \sum_{k=1}^n \begin{bmatrix}
 a_{11k} \boldsymbol{P}_{k:}^T &  a_{12k} \boldsymbol{P}_{k:}^T & \ldots &  a_{iLk} \boldsymbol{P}_{k:}^T \\
\vdots  & \vdots & \ddots & \vdots \\
a_{M1k} \boldsymbol{P}_{k:}^T & a_{M2k} \boldsymbol{P}_{k:}^T & \ldots &  a_{MLk} \boldsymbol{P}_{k:}^T \end{bmatrix}. 
\end{equation*}
Let $\boldsymbol{A}_k \in \R^{M\times L}$, for $k=1,\ldots,n$, be the real-valued matrices such that 
\begin{equation*}
 \matA = \sum_{k=1}^n \boldsymbol{A}_k e_k.
\end{equation*}
In words, $\boldsymbol{A}_k$ is the ``matrix'' component associated with the basis element $e_k$ of $\matA$. 
Using $\boldsymbol{A}_k \in \R^{M\times L}$, we conclude that
\begin{equation} 
\label{eq:kronecker-matrix}
\mathcal{M}_L(\matA) = \sum_{k=1}^n \boldsymbol{A}_k \otimes \boldsymbol{P}_{k:}^T.
\end{equation}
Therefore, $\matC = \matA \matB$ can be efficiently computed by the equation
\begin{equation*} 
\matC = \varphi^{-1} \left(\Big(\sum_{k=1}^n \boldsymbol{A}_k \otimes \boldsymbol{P}_{k:}^T\Big) \varphi(\matB) \right).
\end{equation*}



\begin{example}[Quaternion matrix product]
Consider the quaternion-valued matrix
 \[ \matA = \begin{bmatrix}
    1 + 2\ii & 3\ii + 4\jj & 5\jj + 6\kk\\
    7 + 8\jj & 9 + 10\kk & 11\ii + 12\kk  
            \end{bmatrix} \in \mathbb{Q}^{2 \times 3},\]
 and the column vector 
 \[
 \vetx = \begin{bmatrix}
        1 + 2\ii + 3\jj + 4\kk \\
        5 + 6\ii + 7\jj + 8\kk \\
        9 + 10\ii + 11\jj + 12\kk
         \end{bmatrix} \in \mathbb{Q}^{3 \times 1}.
            \] 
Using quaternion matrix algebra, we obtain
\[ \vety =  \matA \vetx = 
\begin{bmatrix}
 -176 + 45\ii + 96\jj + 11\kk \\
 -306 - 3\ii + 140\jj + 363\kk
\end{bmatrix} \in \mathbb{Q}^{2\times 1}.
\]
To determine the quaternion-valued vector $\vety$ using real-valued matrix computation, we first compute
 \begin{align*}
  \mathcal{M}_L(\matA) &= 
   \begin{bmatrix}
   1 & 0 & 0 \\
   7 & 9 & 0
  \end{bmatrix}\otimes \begin{bmatrix}
  1 & 0 & 0 & 0 \\
  0 & 1 & 0 & 0 \\
  0 & 0 & 1 & 0 \\
  0 & 0 & 0 & 1 \end{bmatrix}+\ldots+
  \begin{bmatrix}
   0 & 0 & 6 \\
   0 & 10 & 12
  \end{bmatrix}\otimes \begin{bmatrix}
  0 & 0 & 0 & -1 \\
  0 & 0 & -1 & 0 \\
  0 & 1 & 0 & 0 \\
  1 & 0 & 0 & 0 \end{bmatrix}
  \\ \setcounter{MaxMatrixCols}{12}
  & = \begin{bmatrix}
    1 &-2 & 0 & 0 & 0 &-3 &-4 & 0 & 0 & 0 &-5 &-6\\
    2 & 1 & 0 & 0 & 3 & 0 & 0 & 4 & 0 & 0 &-6 & 5\\
    0 & 0 & 1 &-2 & 4 & 0 & 0 &-3 & 5 & 6 & 0 & 0\\
    0 & 0 & 2 & 1 & 0 &-4 & 3 & 0 & 6 &-5 & 0 & 0\\
    7 & 0 &-8 & 0 & 9 & 0 & 0& -10 & 0& -11 & 0& -12\\
    0 & 7 & 0 & 8 & 0 & 9& -10 & 0 &11 & 0& -12 & 0\\
    8 & 0 & 7 & 0 & 0 &10 & 9 & 0 & 0 &12 & 0& -11\\
    0 &-8 & 0 & 7 &10 & 0 & 0 & 9 &12 & 0 &11 & 0
  \end{bmatrix}.
 \end{align*} 
Then, we obtain
  \begin{align*}
   \varphi(\vety) &=
    \begin{bmatrix}
    1 &-2 & 0 & 0 & 0 &-3 &-4 & 0 & 0 & 0 &-5 &-6\\
    2 & 1 & 0 & 0 & 3 & 0 & 0 & 4 & 0 & 0 &-6 & 5\\
    0 & 0 & 1 &-2 & 4 & 0 & 0 &-3 & 5 & 6 & 0 & 0\\
    0 & 0 & 2 & 1 & 0 &-4 & 3 & 0 & 6 &-5 & 0 & 0\\
    7 & 0 &-8 & 0 & 9 & 0 & 0& -10 & 0& -11 & 0& -12\\
    0 & 7 & 0 & 8 & 0 & 9& -10 & 0 &11 & 0& -12 & 0\\
    8 & 0 & 7 & 0 & 0 &10 & 9 & 0 & 0 &12 & 0& -11\\
    0 &-8 & 0 & 7 &10 & 0 & 0 & 9 &12 & 0 &11 & 0
  \end{bmatrix} 
  \begin{bmatrix}
    1 \\  2\\  3\\  4\\  5\\  6\\  7\\  8\\  9\\ 10\\ 11\\ 12
  \end{bmatrix}  = \begin{bmatrix}
    -176 \\   45 \\   96\\   11\\ -306\\   -3\\  140\\  363
      \end{bmatrix},
  \end{align*}
which corresponds to a row scan of the elements of the quaternion-valued vector $\vety$.

\end{example}

\section{Vector-Valued Neural Networks (V-Nets)} \label{sec:VvNNs}

Vector-valued neural networks (V-nets) are artificial neural networks conceived to process arrays of vectors. Let us begin addressing dense layers of neurons defined on a finite-dimensional algebra $\mathbb{V}$. 

\subsection{Dense Layers and the Approximation Capability of the Vector-Valued Multilayer Perceptron}

Dense layers, also known as fully connected layers, are the building blocks of several neural network architectures \cite{Geron19HandsOn}. In particular, the famous multilayer perceptron (MLP) network is given by the composition of a sequence of dense layers \cite{Gonzalez2018DeepNetworks}. 

Dense layers are composed of several parallel neurons, each receiving inputs through synaptic connections. Each neuron processes data through a linear combination of its inputs by the synaptic weights (trainable parameters), to which a scalar bias term is added. A non-linear activation function can be applied to yield the neuron's output. In mathematical terms, the output of the $i$th vector-valued neuron in a dense layer is defined by 
\begin{equation*} 
y_i = \psi\left(s_i+b_i\right) \quad \text{with} \quad s_i = \sum_{j=1}^N w_{ij}x_j, \quad \forall i=1,\ldots,M,
\end{equation*}
where $x_1,\ldots,x_N \in \mathbb{V}$ are the vector-valued inputs, $w_{ij} \in \mathbb{V}$ represents the synaptic weight from input $j$ to neuron $i$, $b_i \in \mathbb{V}$ denotes the bias term, and $\psi:\mathbb{V} \to \mathbb{V}$ is a vector-valued activation function. Because $\mathbb{R}$ is a one-dimensional vector space, traditional and vector-valued dense layers are equivalent when $dim(\mathbb{V})=1$. 

\begin{remark}
Because an algebra $\mathbb{V}$ may not be commutative, a dense layer can be alternatively defined as follows:
\begin{equation*} 
y_i = \psi\left(s_i+b_i\right) \quad \text{with} \quad s_i = \sum_{j=1}^N x_j w_{ji}, \quad \forall i=1,\ldots,M.
\end{equation*}
\end{remark}

Choosing the appropriate activation function for an image or signal-processing task can prove to be a challenging issue. To keep things simple, this paper focuses only on the so-called \textit{split activation functions} \cite{Parcollet2020ANetworks,Lee2022Complex-ValuedSurvey,Arena1997MultilayerFunctions}. Briefly, a split activation function is obtained by applying a real-valued function in each coordinate of its vector-valued argument. Formally, we have: 
\begin{definition}[Split Activation Functions] \label{def:split_function}
Let $\basis$ be an ordered basis for a vector space $\mathbb{V}$. A split activation function $\psi: \mathbb{V} \to \mathbb{V}$ is derived from a real-valued function $\psi_\mathbb{R}: \mathbb{R} \to \mathbb{R}$ as follows:
\begin{equation}\label{eq:split_function}
\psi(x) = \sum_{i=1}^n \psi_{\mathbb{R}}(x_i)e_i, \quad \forall x=\sum_{i=1}^n x_i e_i \in \mathbb{V}.
\end{equation}
\end{definition}

\begin{remark}
A split-activation function given by \eqref{eq:split_function} satisfies the identity
\begin{equation} \label{eq:split_property}
    \varphi\big(\psi(x)\big) = \psi_\mathbb{R}\big(\varphi(x)\big),
\end{equation}
where $\varphi$ is the isomorphism defined by \eqref{eq:varphi} and $\psi_\mathbb{R}$ is computed component-wise. Recall that, in practice, $x \in \mathbb{V}$ is identified by $\varphi(x)$. Thus,  \eqref{eq:split_property} is useful as it shows that the representation $\varphi(\psi(x))$ of $\psi(x)$ can be found by evaluating $\psi_\mathbb{R}$ on $\varphi(x)$, which represents $x$.
\end{remark}

The split $\mathtt{relu}$ and split sigmoid functions are examples of vector-valued activation functions used in V-nets, including hypercomplex-valued neural networks \cite{hirose12}.


Despite being computationally expensive due to its numerous parameters, dense layers are widely used because they result in the universal approximation theorem. Briefly, the universal approximation theorem asserts that MLP networks can approximate continuous functions with any desired accuracy on a compact. More specifically, the set of single hidden-layer MLP networks with an appropriate activation function is dense in the set of all continuous functions over a compact. In mathematical terms, the universal approximation theorem for real-valued MLP networks poses conditions on the activation function $\psi:\mathbb{R} \to \mathbb{R}$ such that the set
\begin{equation*}
    \mathcal{H}_{\mathbb{R}} = \left\{ \sum_{i=1}^M \alpha_i \psi\left(\sum_{j=1}^N w_{ij}x_j + b_i\right): M \in \mathbb{N}, \alpha_i,  w_{ij},b_i \in \mathbb{R}  \right\},
\end{equation*}
of all single hidden-layer networks is dense on the set $\mathcal{C}(K)$ of all continuous functions over a compact $K \subseteq \mathbb{R}^N$. Many currently used activation functions like the \texttt{relu} and the logistic functions result in the universal approximation capability. A comprehensive review of the approximation capability of traditional MLP networks can be found in \cite{Pinkus1999ApproximationNetworks}. 

Recently, many researchers addressed the approximation capabilities of neural networks, including deep and shallow models based on piece-wise linear activation functions \cite{Petersen2018OptimalNetworks}. Moreover, the approximation capability of hypercomplex-valued neural networks has been investigated by several researchers, including Arena et al., \cite{Arena1997MultilayerFunctions}, Buchholz and Sommer \cite{Buchholz2000HyperbolicPerceptron,Buchholz2001CliffordPerceptrons}, and more recently Voigtlaender \cite{Voigtlaender2023TheNetworks}. 
More generally, based on Arena et al. \cite{Arena1997MultilayerFunctions} and Vital et al. \cite{Vital2022ExtendingNetworks}, we can state the following theorem concerning vector-valued multilayer perceptron (V-MLP) networks: 
\begin{theorem}[Universal Approximation Theorem for V-MLP Networks] \label{thm:approximation}
Let $\psi_\mathbb{R}: \mathbb{R} \to \mathbb{R}$ be a real-valued activation function that yields approximation capability to the set $\mathcal{H}_\mathbb{R}$ of all single hidden-layer MLP networks and satisfies $\lim_{t  \to -\infty} \psi_{\mathbb{R}}(t)=0$. Consider a finite-dimensional non-degenerate algebra $\mathbb{V}$,  let $\psi:\mathbb{V} \to \mathbb{V}$ be the split activation function derived from $\psi_\mathbb{R}$, and let $K \subset \mathbb{V}^N$ be a compact set. 
The class of single hidden-layer V-MLP networks with real-valued output weights given by 
\begin{equation}\label{eq:class_V-MLP}
    \mathcal{H}_{\mathbb{V}} = \left\{ \sum_{i=1}^M \alpha_i \psi\left(\sum_{j=1}^N w_{ij}x_j + b_i\right): M \in \mathbb{N}, \alpha_i \in \mathbb{R},  w_{ij},b_i \in \mathbb{V}  \right\},
\end{equation}
is dense in the set $\mathcal{C}(K)$ of all continuous functions from $K$ to $\mathbb{V}$. Furthermore, if $\mathbb{V} \equiv \mathbb{H}$ is a hypercomplex algebra, then the class of single hidden-layer hypercomplex-valued MLP networks given by 
\begin{equation}\label{eq:class_H-MLP}
    \mathcal{H}_{\mathbb{H}} = \left\{ \sum_{i=1}^M \alpha_i \psi\left(\sum_{j=1}^N w_{ij}x_j + b_i\right): M \in \mathbb{N}, \alpha_i \in \mathbb{H},  w_{ij},b_i \in \mathbb{H}  \right\},
\end{equation}
is dense in the set $\mathcal{C}(K)$.
\end{theorem}
Theorem \ref{thm:approximation} says that V-MLP networks with split activation function inherit the approximation capability of traditional models if the algebra is non-degenerate. Furthermore, the split activation function must be derived from an activation function $\psi_\mathbb{R}$ such that $\lim_{t  \to -\infty} \psi_{\mathbb{R}}(t)=0$. In this case, given a continuous function $f:K \to \mathbb{V}$ and $\epsilon>0$, Theorem \ref{thm:approximation} ensures that there exists a shallow V-MLP network $\mathcal{N}:\mathbb{V}^N \to \mathbb{V}$, with real-valued output weights, such that
\begin{equation}
    | f(\boldsymbol{x})-\mathcal{N}(\boldsymbol{x})|< \epsilon, \quad \forall \boldsymbol{x} \in K.
\end{equation}
Note that the V-MLP network maps $\mathbb{V}^N$ into $\mathbb{V}$ despite the output weights being real numbers. Moreover, the real-valued output weights can be replaced by vector-valued ones if $\mathbb{V}$ is a hypercomplex algebra. In other words, a V-MLP of exclusively vector-valued dense layers has the universal approximation property if $\mathbb{V}$ is a finite-dimensional non-degenerate algebra with identity.

\subsection{Relationship Between Real and Vector-Valued Dense Layers}  \label{sec:VvNN2Real}

This subsection discusses the relationship between traditional and vector-valued dense layers. To simplify the exposition, we formulate a dense layer using matrix operations.  

From a computational point of view, dense layers are efficiently implemented using matrix and vector operations. Accordingly, the output $\boldsymbol{y} = (y_1,\ldots,y_M) \in \mathbb{V}^M$ of a dense layer with $M$ vector-valued neurons in parallel is determined by the equation
\begin{equation}
    \label{eq:dense_layer}
    \boldsymbol{y} = \boldsymbol{\psi}(\boldsymbol{s} +\boldsymbol{b}) \quad \text{with} \quad \boldsymbol{s} = \boldsymbol{W}\boldsymbol{x},
\end{equation}
where $\boldsymbol{x} = (x_1,\ldots,x_N) \in \mathbb{V}^N$ is the input vector, $\boldsymbol{W} = (w_{ij}) \in \mathbb{V}^{M\times N}$ is the matrix containing the synaptic weights, $\boldsymbol{b}=(b_1,\ldots,b_M) \in \mathbb{V}^M$ is the bias vector, and $\boldsymbol{\psi}:\mathbb{V}^M \to \mathbb{V}^M$ is defined in a component-wise manner by means of the following equation for some $\psi:\mathbb{V} \to \mathbb{V}$: 
\begin{equation*}
    [\boldsymbol{\psi}(\boldsymbol{s})]_i = \psi(s_i), \quad \forall i=1,\ldots,M.
\end{equation*}

In practice, we usually work with the real-valued representation of the inputs and outputs because current deep-learning libraries operate almost exclusively with floating-point numbers. Precisely, using the isomorphism $\varphi$ given by \eqref{eq:Matrix-varphi}, we consider $\varphi(\vetx) \in \mathbb{R}^{nN}$ and $\varphi(\vety)\in \mathbb{R}^{nM}$ instead of $\vetx \in \mathbb{V}^N$ and $\vety \in \mathbb{V}^M$, respectively. Now, from \eqref{eq:Matrix-vec-prod} and \eqref{eq:split_property}, a vector-valued dense layer given by \eqref{eq:dense_layer} can be emulated by a real-valued dense layer defined by
\begin{equation}
    \label{eq:dense_hyper2real}
    \varphi(\boldsymbol{y}) = \boldsymbol{\psi}_{\mathbb{R}}\big(\varphi(\boldsymbol{s}) + \varphi(\boldsymbol{b})\big) \quad \text{with} \quad \varphi(\boldsymbol{s}) = \mathcal{M}_L(\boldsymbol{W})\varphi(\boldsymbol{x}),
\end{equation}
where $\varphi(\vetx) \in \mathbb{R}^{nN}$ is a real-valued input vector, $\mathcal{M}_L(\boldsymbol{W}) \in \mathbb{R}^{nM \times nN}$ is a real-valued synaptic weight matrix, $\varphi(\boldsymbol{b})\in \mathbb{R}^{nM}$ is a real-valued bias vector, and $\varphi(\vety) \in \mathbb{R}^{nM}$ is the real-valued output. In other words, a vector-valued dense layer can be computed using a real-valued dense layer by appropriately rearranging the elements of the vector-valued arrays. In particular, from \eqref{eq:kronecker-matrix}, we have
\begin{equation*}
    \mathcal{M}_L(\boldsymbol{W}) = \sum_{k=1}^n \boldsymbol{W}_k \otimes \boldsymbol{P}_{k:}^T,
\end{equation*}
where the vector-value matrix is written as $\boldsymbol{W} = \boldsymbol{W}_1 e_2 + \ldots + \boldsymbol{W}_n e_n$ using the basis $\basis$ and the matrices $\boldsymbol{P}_{1:}^T,\ldots,\boldsymbol{P}_{:n}^T$ depends on the algebra. Therefore, the real-valued matrix $\mathcal{M}_L(\boldsymbol{W})$ associated with a vector-valued dense layer has $n(MN)$ distinct trainable parameters. In contrast, the synaptic weight matrix of a traditional dense layer has $n^2MN$ trainable parameters. Including the bias vector, we conclude that a vector-valued dense layer has $nM(N+1)$ parameters while an equivalent traditional dense layer has $nM(nN+1)$ parameters. Thus, vector-valued dense layers can be interpreted as constrained versions of traditional dense layers, where the synaptic weights are obtained by imposing a structure that depends on the algebra. Furthermore, the constraints imposed by the algebra follow from supposing that intercorrelations exist between feature channels.

Finally, in certain applications, the network may require the output of real-valued vectors instead of vector-valued arrays, even though the input is vector-valued because of the intercorrelation between feature channels. For example, only the real part of the output of hypercomplex-valued neural networks has been used for acute lymphoblastic leukemia detection through blood smear digital microscopic images in \cite{Vieira2022AcuteNetworks}. One can handle such scenarios by focusing on a single component of a vector-valued output. Precisely, the output of a vector-valued dense layer can be written as $\vety = \sum_{k=1}^n \vety_k e_k \in \mathbb{V}^M$, where $\vety_1, \ldots, \vety_n$ are all real-vectors in $\mathbb{R}^M$. Thus, one can consider the component $\vety_k \in \mathbb{R}^M$, for some $k \in \{1,\ldots,n\}$, as the real-valued output of the network. However, the output component $\vety_k$ of a vector-valued dense layer with split activation function is equivalent to the output of a real-valued dense layer applied on a flattened version of the input. Indeed, using a split activation function, we conclude from \eqref{eq:dense_layer} that $\vety_{k} = \psi_{\mathbb{R}}(\boldsymbol{s}_k + \boldsymbol{b}_k)$, where $\boldsymbol{s} = \sum_{k=1}^n \boldsymbol{s}_k e_k$ and $\boldsymbol{b} = \sum_{k=1}^n \boldsymbol{b}_k e_k$. Now, let $s_{ki}$ denote the $i$th entry of $\boldsymbol{s}_k$. From Proposition \ref{prop:bilinear}, we have 
\begin{equation} \label{eq:s_ki}
    s_{ki} = \sum_{j=1}^N \mathcal{B}_k(w_{ij},x_j)
    = \sum_{j=1}^N \varphi(w_{ij})^T \boldsymbol{B}_k \varphi(x_j)
    = \sum_{j=1}^N \hat{\boldsymbol{w}}_{ij} \varphi(x_j), \quad \forall i=1,\ldots,M,
\end{equation}
where $\hat{\boldsymbol{w}}_{ij} = \varphi(w_{ij})^T \boldsymbol{B}_k \in \mathbb{R}^{1 \times n}$ is a row vector for all $i$ and $j$. Using matrix notation, \eqref{eq:s_ki} can be written as $\boldsymbol{s}_k = \hat{\boldsymbol{W}} \varphi(\vetx)$ where
\begin{equation*}
    \hat{\boldsymbol{W}} = \begin{bmatrix}
        \hat{\boldsymbol{w}}_{11} & \ldots & \hat{\boldsymbol{w}}_{1N} \\
        \vdots & \ddots & \vdots \\
        \hat{\boldsymbol{w}}_{M1} & \ldots & \hat{\boldsymbol{w}}_{MN}
    \end{bmatrix} \in \mathbb{R}^{M \times (nN)}.
\end{equation*}
Concluding, the $k$th component of the vector-valued output $\vety \in \mathbb{V}^M$ satisfies $\vety_k = \psi_{\mathbb{R}}(\hat{\boldsymbol{W}} \varphi(\vetx)+\boldsymbol{b}_k)$, which means $\vety_k$ is the output of a real-valued dense layer computed on the flattened version $\varphi(\vetx) \in \mathbb{R}^{nN}$ of the input $\vetx \in \mathbb{V}^N$.


\subsection{Vector-valued Convolutions and Some Remarks on Other Building Blocks}

Convolutional layers are important building blocks in current deep learning models. Broadly speaking, convolutional layers are special layers in which the units are connected to small sections in the feature maps of the preceding layer through a set of weights called a filter bank \cite{Lecun2015DeepLearning}. Moreover, all units share the same filter banks. Besides reducing significantly the number of parameters, convolutional layers exhibit spatial invariance and are effective for the detection of local patterns. 

A vector-valued convolutional layer is defined as follows. Let $\mathbf{\vetx}$ be the input (image or signal) with $C$ feature channels. We denote by $\vetx(p,c) \in \mathbb{V}$ the content of $\vetx$ in the $c$th channel at location $p \in D_{\vetx}$, where $\mathcal{D}_{\vetx}$ denotes the domain of $\vetx$. The weights of a
convolutional layer with $K$ filters are arranged in an array $\mathbf{W}$ such that $\mathbf{W}(q,c,k) \in \mathbb{V}$ corresponds to the weight of the $k$th filter in the $c$th channel at $q \in D$, where $D$ denotes the filters' domain\footnote{The domain $D$ is usually a rectangular grid in image processing tasks.}.
The vector-valued convolution of $\mathbf{W}$ and $\vetx$, denoted by $\mathbf{W} \ast \vetx$, is given by the sum of the cross-correlation of $\mathbf{W}(:,c,k)$ and $\vetx(:,c)$ over all channels $c=1,\ldots,C$. Precisely, we have
\begin{equation}
   \label{eq:conv}
   (\mathbf{W} \ast \vetx)(p,k) = \sum_{c=1}^C \sum_{q \in D}\mathbf{W}(q,c,k) \vetx(p+S(q),c), \quad p \in \mathcal{D}_{\vety},\;\forall k=1,\ldots,K,
\end{equation}
where $p+S(q)$ denotes a translation that can take strides into account and $\mathcal{D}_{\vety}$ denotes the domain of $\vety$. The output $\vety$ of a convolutional layer is obtained by evaluating an activation function on the addition of a bias term $\boldsymbol{b}$ with the convolution $\mathbf{W} \ast \vetx$. In mathematical terms, we have 
$\vety = \psi(\mathbf{W}\ast \vetx + \boldsymbol{b})$, where the activation function $\psi:\mathbb{V} \to \mathbb{V}$ is applied in an entry-wise manner.

We would like to remark that a traditional convolutional layer is obtained when $\mathbf{W}(q,c,k)$ and $\vetx(p,c)$ are real numbers instead of vectors. In this case, the convolution $\mathbf{W} \ast \vetx$ is obtained by summing over all the real-valued feature channels. In contrast, the vector-valued convolution operates under the assumption that there is an intercorrelation between $n$ feature channels, which is achieved through the use of vector multiplication in \eqref{eq:conv}. 

Like dense layers, vector-valued convolutional layers can be emulated using real-valued convolutional layers, which is particularly useful for implementing convolutional neural networks using current deep learning libraries. Accordingly, using the isomorphism $\varphi:\mathbb{V} \to \mathbb{R}^n$, the Kronecker product, and the linearity of the convolution operation, we have
\begin{align}
\label{eq:conv2real}
    \varphi\big(\mathbf{W} \ast \vetx \big)
    = \left(\sum_{\ell=1}^n \mathbf{W}_\ell \otimes \mathbf{P}_{\ell:}^T\right) \ast \varphi(\vetx)
    = \sum_{\ell=1}^n \left(\mathbf{M}_\ell \ast  \varphi(\vetx) \right),
\end{align}
where $\mathbf{W} =\mathbf{W}_1 e_1+\ldots+\mathbf{W}_n e_n$ is the representation of the filters with respect to the basis $\basis$, $\mathbf{M}_\ell = \mathbf{W}_\ell \otimes \mathbf{P}_{\ell:}^T$ are real-valued filters obtained using the Kronecker product, and $\varphi(\vetx)$ is obtained concatenating the components $\vetx_1,\ldots,\vetx_n$ of $\vetx = \vetx_1e_1+\ldots+\vetx_n e_n$ into the feature axis. Examples of the vector-valued convolutions and their implementation can be found in \cite{Vieira2022AcuteNetworks,Grassucci2022PHNNs:Convolutions,Gaudet2018DeepNetworks}. We would like to remark that split activation functions are particularly helpful in the emulation of vector-valued convolutional layers by real-valued ones. Indeed, from \eqref{eq:split_property}, the concatenation $\varphi(\vety)$ of the output $\vety$ is obtained by evaluating a real-valued activation function $\psi_{\mathbb{R}}$ entry-wise in the sum $\varphi(\mathbf{W}\ast \vetx) + \varphi(\boldsymbol{b})$, with $\varphi(\mathbf{W}\ast \vetx)$ given by \eqref{eq:conv2real}.

In addition to convolutional layers, modern deep learning models utilize other structures, such as pooling layers and batch normalization \cite{Geron19HandsOn}. While vector-valued versions of these structures are a topic of future research, we can currently use a simpler approach. This approach involves combining traditional structures with real-valued emulation of vector-valued convolutional and dense layers, as described in \eqref{eq:dense_hyper2real} and \eqref{eq:conv2real}. Although this approach may seem overly simplistic, it can still provide valuable insights into the vector-valued blocks. For example, when a max-pooling layer is applied to the real-valued representation of a vector-valued image, the result is equivalent to computing the max-pooling with the maximum given by the so-called marginal or Cartesian product ordering of vectors. Additionally, this pooling layer corresponds to extending the pooling layer in a split manner, as shown in equation \eqref{eq:split_function}.

\section{Concluding Remarks} \label{sec:concluding}

Despite the many successful applications of traditional neural networks for signal and image processing tasks, they do not initially take into account the intercorrelation between feature channels. Such intercorrelation is learned from the dataset, but it demands careful consideration when selecting optimization methods and hyperparameters, as well as using appropriate regularization strategies. In contrast, vector-valued neural networks (V-nets) naturally incorporate the intercorrelation between feature channels through the vector algebra. Furthermore, using hypercomplex algebras can lead to additional geometric and algebraic properties in the network model \cite{Ruhe2023GeometricNetworks,Parcollet2020ANetworks,hirose12,Lee2022Complex-ValuedSurvey}. {Recent successful applications of V-nets, predominantly hypercomplex-valued neural networks, include sound event localization and detection \cite{Grassucci2023DualRepresentation}, improving ultrasound image quality \cite{Lei2023FullyImaging}, acute lymphoblastic leukemia detection \cite{Vieira2022AcuteNetworks}, large-scale fluid dynamics simulations \cite{Ruhe2023GeometricNetworks}, to list a few. It is worth noting that V-nets outperformed traditional neural networks in all of the aforementioned applications.}

This paper provided the basic concepts of V-nets. We began reviewing the concept of algebra, which is obtained by enriching a vector space with a multiplication. We defined a hypercomplex algebra as an algebra with an identity. Therefore, hypercomplex-valued neural networks are V-nets with additional geometric or algebraic properties. In particular, from Example \ref{ex:PHA}, we conclude that the recently introduced parameterized ``hypercomplex'' neural networks are, in fact, V-nets \cite{Grassucci2022PHNNs:Convolutions,Zhang2021BeyondParameters}.
In addition, this paper establishes the relationship between V-nets and traditional neural networks. Precisely, we showed how vector-valued dense and convolutional layers can be emulated using real-valued ones. Such emulation is particularly helpful for the implementation of V-nets using current deep-learning libraries like \texttt{tensorflow} and \texttt{pytorch}. Besides making the implementation of V-nets straightforward, we can utilize automatic differentiating features to train V-nets without having to deal with complicated vector or hypercomplex calculus for the gradients. Moreover, other traditional building blocks can be combined with vector-valued convolutional and dense. Such a straightforward approach yielded promising results in image and signal processing applications \cite{Vieira2022AcuteNetworks,Grassucci2022PHNNs:Convolutions,Parcollet2020ANetworks,Lei2023FullyImaging}.
Besides the practical considerations, this paper also addressed important theoretical issues. Namely, Theorem \ref{thm:approximation} concerns the approximation capability of vector-valued multilayer perceptron defined on finite-dimensional non-degenerate algebras. 

Concluding, V-nets are obtained by imposing certain intercorrelation between the features. As a consequence, they provide a graceful approach to the bias-variance trade-off by incorporating into the neural network's operation additional characteristics of the data. Future research should focus on selecting the appropriate algebra for image and signal processing applications. Efficient techniques for learning the most suited algebra for a task also require further research.  Furthermore, vector-valued activation functions beyond the split functions, as well as deep-learning building blocks and tools like batch normalization and dropout strategies, should be further investigated in the future. 

\section*{Acknowledgment}
\addcontentsline{toc}{section}{Acknowledgment}

I would like to acknowledge that this paper is partially based on a minicourse I held at the Institute for Research and Applications of Fuzzy Modeling, University of Ostrava in the Czech Republic in February 2023. This minicourse was part of the Strategic Research Partnership called "Mathematical Aspects of Complex, Hypercomplex, and Fuzzy Neural Networks," which was supported by the Polish National Agency for Academic Exchange. I am grateful to Agnieszka Niemczynowicz, the partnership coordinator, and the project collaborators for their valuable insights and discussions.

\end{document}